\documentclass[11pt]{article}

\usepackage[preprint]{acl}

\usepackage{times}
\usepackage{latexsym}
\usepackage{amsmath}
\usepackage{array}   
\usepackage{makecell} 
\usepackage{enumitem}
\usepackage{algorithm}
\usepackage{float} 
\usepackage{multirow}
\usepackage{booktabs}
\usepackage{stfloats}
\usepackage{placeins}
\usepackage{algpseudocode}

\usepackage[T1]{fontenc}

\usepackage[utf8]{inputenc}

\usepackage{microtype}

\usepackage{inconsolata}

\usepackage{graphicx}

\title{A Semantic Decoupling–Based Two-Stage Rainy-Day Attack for Revealing Weather Robustness Deficiencies in Vision–Language Models
}

\author{
Chengyin Hu \enspace
Xiang Chen \enspace
Zhe Jia \enspace
Weiwen Shi \enspace
Fengyu Zhang \enspace
Jiujiang Guo \enspace
Yiwei Wei
}

\begin{document}
\maketitle
\begin{abstract}
Vision–Language Models (VLMs) are trained on image–text pairs collected under canonical visual conditions and achieve strong performance on multimodal tasks. However, their robustness to real–world weather conditions, and the stability of cross–modal semantic alignment under such structured perturbations, remain insufficiently studied. In this paper, we focus on rainy scenarios and introduce the first adversarial framework that exploits realistic weather to attack VLMs, using a two-stage, parameterized perturbation model based on semantic decoupling to analyze rain-induced shifts in decision-making. In Stage 1, we model the global effects of rainfall by applying a low-dimensional global modulation to condition the embedding space and gradually weaken the original semantic decision boundaries. In Stage 2, we introduce structured rain variations by explicitly modeling multi–scale raindrop appearance and rainfall–induced illumination changes, and optimize the resulting non-differentiable weather space to induce stable semantic shifts. Operating in a non–pixel parameter space, our framework generates perturbations that are both physically grounded and interpretable. Experiments across multiple tasks show that even physically plausible, highly constrained weather perturbations can induce substantial semantic misalignment in mainstream VLMs, posing potential safety and reliability risks in real-world deployment. Ablations further confirm that illumination modeling and multi-scale raindrop structures are key drivers of these semantic shifts.

\end{abstract}
\section{Introduction}

Vision–Language Models (VLMs) achieve strong performance by aligning visual and linguistic representations through large-scale joint pre-training \citep{ref14, jia2021noisytext, li2021momentumdistill, li2022blip}. By grounding images in high-level semantic concepts, VLMs enable unified cross-modal reasoning and consistently surpass vision-only models in image classification, captioning, and visual question answering (VQA) \citep{ref1, ref2, ref3, ref4, ref5}. Unlike conventional CNNs relying on local pixel statistics \citep{ref34, ref35, ref37}, VLMs interpret images through semantic abstraction, enhancing generalization across tasks and environments.

\begin{figure}[t]
    \centering
    \includegraphics[width=1\linewidth]{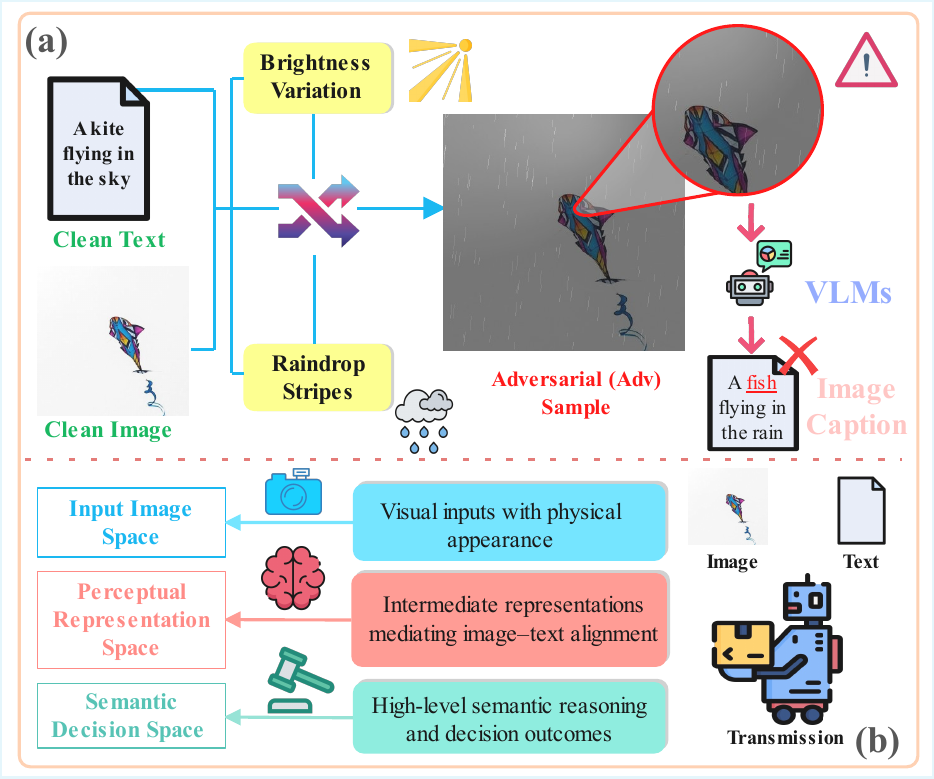}
    \caption{(a) Raindrop streaks and brightness variations disrupt correlated appearance cues, leading to a misalignment between vision and language in image captioning. (b) Vision–language reasoning is decoupled into input, representation, and semantic decision spaces.}
    \label{fig:framework}
    \vspace{-0.5cm}
\end{figure}

However, this semantic-centered design introduces a unique vulnerability. Semantic representations in VLMs depend on the global coherence of visual cues such as shape, color, brightness, texture, and spatial layout \citep{refa, refb, refc}. When these cues are systematically disturbed by changes in appearance or illumination, image–text alignment can deteriorate even if objects remain visually recognizable. Unlike pixel-level noise \citep{ref000, ref001}, such structured perturbations directly distort semantic representations and impair VLM reasoning. Real-world weather conditions such as rain, fog, and snow naturally cause correlated disruptions through interacting physical factors. As shown in Figure~\ref{fig:framework}(a), in image captioning, weather-induced perturbations jointly distort visual appearance and contextual cues, causing misalignment between images and text by corrupting the semantic grounding of the primary subject.

Despite the ubiquity of such conditions in real-world environments, existing robustness evaluations of VLMs mainly target synthetic noise or simple, isolated corruptions, such as uniform brightness shifts or Gaussian noise \citep{ref38, ref39, zhao2023robustvlm, ref12}. These settings fail to reflect the compound nature of real weather, where multiple physical factors jointly shape semantic perception. Consequently, current benchmarks likely underestimate the vulnerability of VLMs in complex environments \citep{ref002, ref003}.

Motivated by this gap, we study the semantic robustness of VLMs under realistic rain from an adversarial perspective. Instead of passive testing, we employ adversarial probing to reveal how complex weather disrupts image–text alignment beyond synthetic or pixel-level perturbations. Directly optimizing physical weather parameters is difficult due to the strong coupling between visual and linguistic representations in VLMs, which often yields shallow appearance changes without meaningful semantic shifts \citep{ref01, ref02}. To overcome this, we propose a two-stage Semantic Decoupling framework that separates semantic manipulation from physical appearance synthesis, as illustrated in Figure~\ref{fig:framework}(b). The framework decomposes vision–language reasoning into the input image, perceptual representation, and semantic decision spaces, exposing vulnerabilities at the perceptual level before realizing them through physically grounded weather synthesis.

Specifically, our framework consists of two stages. In \textbf{Stage 1}, we modulate the image–text embedding space to weaken the semantic alignment learned by VLMs. This design is motivated by the observation that realistic rainy conditions primarily affect high-level semantics rather than isolated pixels. By relaxing semantic boundaries in advance, subsequent weather-induced effects can more effectively induce semantic misalignment. In \textbf{Stage 2}, we optimize a parameterized rainy-weather space that jointly models multi-scale raindrops and spatial illumination. Using CMA-ES \citep{ref13}, we generate physically plausible rainy patterns that exploit the loosened semantic structure to maximize semantic misalignment.
Unlike approaches that accumulate pixel-level noise, our framework provides an interpretable and physically grounded perspective on how different weather components contribute to semantic degradation in VLMs. Compared to prior work, our method produces realistic perturbations, improves optimization stability through semantic decoupling, and demonstrates strong transferability across tasks and model architectures.
Our contributions are summarized as follows:
\begin{itemize}[left=0em, itemsep=0pt, topsep=4pt, labelsep=1em]
  \item We are the first to systematically study the robustness of VLMs under rainy conditions, with a focus on semantic degradation rather than pixel-level distortions.
  \item We propose a two-stage framework based on semantic decoupling that aligns with the semantic reasoning mechanism of VLMs and the compound nature of weather effects.
  \item We evaluate our approach on zero-shot image classification, image captioning, and visual question answering, and provide detailed analyses of semantic shifts.
\end{itemize}

\section{Related Work}

\textbf{Text-side Manipulation and Sensitivity Analysis.} Text-side variations provide an alternative pathway for influencing vision–language model inference by manipulating linguistic inputs without modifying visual content. Changes in descriptions, prompt templates, or phrasing can substantially affect model predictions, as evidenced by the strong prompt sensitivity observed in CLIP-style zero-shot classification \citep{ref14}. Subsequent prompt-learning approaches introduce learnable textual prototypes or contextual vectors, further emphasizing the dominant role of language in shaping cross-modal alignment \citep{ref15, ref16}. Beyond performance effects, recent studies demonstrate that adversarially crafted prompts can steer model behavior or induce unintended and potentially unsafe outputs without altering the visual input \citep{ref17, ref18}.

\noindent\textbf{Visually Grounded Adversarial Attacks without Physical Constraints.}\quad Small-magnitude pixel-space perturbations are widely used to attack VLMs by disrupting cross-modal alignment. By transferring CNN-based adversarial techniques to CLIP-style architectures, prior studies show that visual-only perturbations can substantially degrade cross-modal matching and downstream performance \citep{ref10, ref19}. Recent analyses \citep{ref11, ref20} further reveal that such perturbations distort joint embedding geometry and erode decision boundaries, exposing representation-level instability as a key source of adversarial vulnerability in VLMs.

\noindent\textbf{Physically Consistent Environmental Perturbations with Semantic-level Analysis.}\quad To improve real-world relevance, recent adversarial evaluations \citep{ref12,ref100} increasingly enforce physical consistency by perturbing interpretable environmental factors rather than unconstrained pixels. ITA \citep{ref12} parameterizes illumination and uses black-box optimization to find adversarial relighting that disrupts cross-modal alignment while preserving perceptual naturalness. Similarly, AdvDreamer \citep{ref100} studies physically grounded adversarial 3D variations, showing that realistic environmental changes can reliably induce semantic failures in VLMs.

\begin{figure*}[h]
    \centering
    \includegraphics[width=1\textwidth]{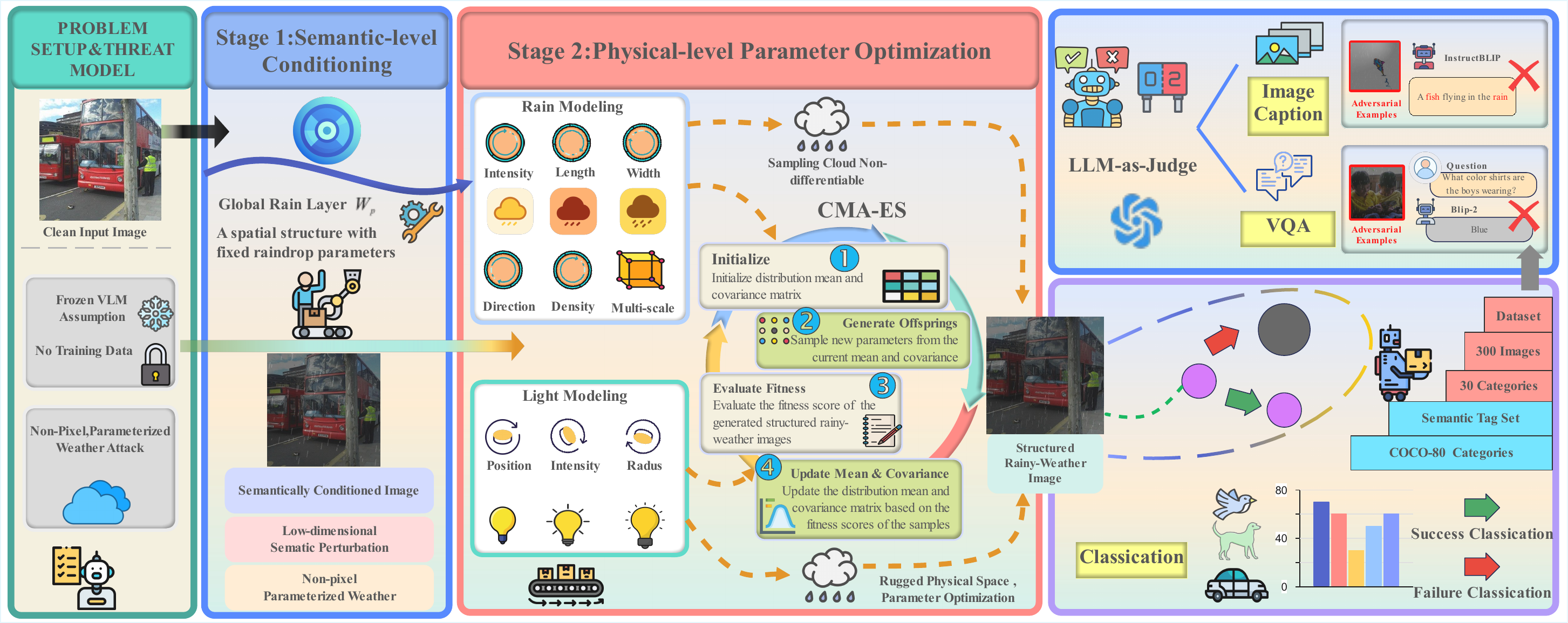}
    \caption{Framework of our method. Stage~1 conditions the embedding space with a global rain layer; Stage~2 uses CMA-ES to optimize parameterized raindrops and illumination to generate structured adversarial rainy images, evaluated on classification and transferred to captioning and VQA.}
    \label{fig2}
    \vspace{-0.3cm}
\end{figure*}

\section{Approach}
\subsection{Framework Overview}
Figure~\ref{fig2} presents a two-stage, semantics-decoupled rainy-weather attack. Stage 1 conditions the cross-modal embedding space via low-dimensional rain-layer mixing, while Stage 2 employs CMA-ES to optimize multi-scale raindrop and illumination parameters for stable semantic shifts. The generated images are evaluated on zero-shot classification and transferred to captioning and VQA using an LLM-as-Judge \citep{bavaresco2025llms}.

\subsection{Model Setup and Problem Definition}

This section explores the problem of robust adversarial attacks on VLMs under real-world physical consistency constraints. The models considered consist of an image encoder and $f_v(\cdot)$ and a text encoder $f_t(\cdot)$, with semantic discrimination achieved through cross-modal similarity. 

In the evaluation, we primarily use a text collection consisting of COCO-80 semantic labels for zero-shot image classification, while reusing the same visual embedding representation for image captioning and visual question answering tasks. Since the attack directly targets the embedding space output by the image encoder, the method itself is independent of the specific task format. Given an input image $I$, the similarity between the image and the text prompt $T_k$ corresponding to the semantic label $K$ is defined as:
\begin{equation}
S_k(I) = \langle f_v(I), f_t(T_k) \rangle
\end{equation}
The model's prediction is determined by the semantic label with the highest similarity.

\subsection{Physical Perturbation Modeling}

\textbf{Parametric Modeling of Raindrop Patterns.}\quad In contrast to pixel-level perturbations, we model raindrops using explicit physical parameters, yielding physically interpretable perturbations. Specifically, raindrop generation is governed by six optimizable variables: intensity $\alpha_{r}$, controlling raindrop brightness contribution; density $\rho_{r}$, determining the number of raindrops per unit area; length $l_{r}$ and width $b_{r}$, capturing motion-induced streaking effects; direction $\varphi_{r}$, simulating wind-driven rainfall; and blur kernel size, approximating scale-dependent blur caused by depth and focus variations. These parameters are collectively represented as a raindrop control vector, defined as:

{
\begin{equation}
\theta_r=(\alpha_r,\rho_r,l_r,b_r,\varphi_r,k_r)
\end{equation}
}

To capture heterogeneous raindrop scales observed in real rainy conditions, we introduce a multi-scale superposition mechanism:
\begin{equation}
R(\theta_r)=\sum_{s\in\mathcal{S}}R_s(\theta_r)
\end{equation}
Here, $R(\theta_r)$ denotes the overall rain perturbation layer, constructed by linearly superposing scale-specific raindrop layers $R_{s}(\cdot)$ over a set of scales $S$. This multi-scale design enables the perturbation to simultaneously influence fine-grained local textures and mid-scale structures, thereby more effectively modulating visual feature extraction.

\noindent\textbf{Illumination Modulation Model.}\quad Considering that real rainfall is often accompanied by localized or non-uniform illumination variations, we introduce an illumination field composed of multiple local light sources:
\begin{equation}
{\small
L(x,y)=\sum_{i=1}^{N} \alpha_{i} \exp\!\left(-\frac{(x-x_i)^2+(y-y_i)^2}{2r_i^2}\right)
}
\end{equation}
Here, $(x_i,y_i)$, $\alpha_{i}$, and $r_i$ denote the position, intensity, and range of the $i$-th light source, respectively. 

The illumination perturbation is applied multiplicatively to image brightness, where $\odot$ denotes element-wise multiplication across spatial locations and color channels. Specifically, the illumination gain map is defined as:
\begin{equation}
G = w_{\ell} L + (1 - w_{\ell}),
\end{equation}
where $L$ denotes the illumination map and $w_{\ell} \in [0,1]$ controls the modulation strength. The modulated image is then obtained by:
\begin{equation}
I_{\mathit{out}} = I_{\mathit{in}} \odot G.
\end{equation}
This illumination modulation is physically grounded in real imaging pipelines and is naturally compatible with the raindrop perturbation.  To characterize global and local illumination variations, we further derive the spatial average $\overline{G}$ and the maximum value $G_{\max}$ of the gain map $G$, together with a reference level $G_0$ and an upper-bound threshold $G_{\mathit{thr}}$. Deviations of $\overline{G}$ from $G_0$ are constrained by a global illumination regularizer weighted by $\lambda_{\mathit{light}}^{(2)}$, while excessive local amplification beyond $G_{\mathit{thr}}$ is penalized by a range regularizer weighted by $\lambda_{\mathit{range}}^{(2)}$.

\subsection{Stage 1: Global Raindrop Layer Mixing}

In the first stage, we fix the spatial structure of the raindrops and control the perturbation intensity by adjusting the mixing weight between the perturbation and the original image. This allows us to rapidly compress the semantic discrimination margin within a low-dimensional parameter space. The adversarial samples in Stage 1 are generated through linear mixing as follows:
\begin{equation}
I^{(1)}=(1-w_p)I+w_pR_{fix}
\end{equation}
Here, $w_{p}$ is the optimizable global raindrop mixing weight that controls raindrop visibility. $R_{fix}$ denotes the fixed global rain layer used in Stage 1.

For the synthesized image  $I^{(1)}$, we define the attack objective as the similarity difference between the real semantic and the most competitive non-real semantic:
\begin{equation}
\mathcal{L}^{(1)}_{\mathit{atk}}
= S_y\big(I^{(1)}\big) - \max_{k\ne y} S_k\big(I^{(1)}\big)
\end{equation}
Minimizing $\mathcal{L}_{\mathrm{\mathit{atk}}}^{(1)}$ progressively reduces the margin between the similarity to the ground-truth text $T_y$ and the strongest competing text $T_k$ in the joint embedding space, thereby eroding the discriminative advantage of the true semantics and moving the sample toward the semantic decision boundary.

To prevent excessive mixing weights from causing significant structural damage, Stage 1 explicitly introduces VGG perceptual consistency~\cite{ref23}, where $\varphi_{l}(\cdot)$ represents the feature response of the pretrained VGG network at the  $l$-th layer. 
\begin{equation}
\mathcal{L}_{\mathit{perc}}^{(1)}=\sum_{\ell\in\mathcal{L}}\left\|\phi_\ell(I^{(1)})-\phi_\ell(I)\right\|_2^2
\end{equation}

Additionally, quadratic regularization is applied to the mixing weights to ensure the stability of the optimization process:
\begin{equation}
\mathcal{L}_{reg}=(w_p-w_p^0)^2
\end{equation}

The overall objective for Stage 1 is as follows:
\begin{equation}
\mathcal{L}_{stage1}=\mathcal{L}_{atk}^{(1)}+\lambda_{perc}^{(1)}\mathcal{L}_{perc}^{(1)}+\lambda_{reg_{}}^{(1)}\mathcal{L}_{reg}
\end{equation}
Here, $\lambda_{perc}^{(1)}$ and $\lambda_{reg}^{(1)}$ control the strengths of the perceptual and regularization terms, respectively.

\begin{table*}[b]
\centering
\caption{Classification accuracy (\%) of different CLIP models under various attack methods.}
\label{tab:1}

\resizebox{\linewidth}{!}{
\begin{tabular}{ccccc}
\toprule
Method
& OpenCLIP ViT-B/16
& Meta-CLIP ViT-L/14
& EVA-CLIP ViT-G/14
& OpenAI CLIP ViT-L/14 \\
\midrule
Clean
& 97
& 98
& 98
& 93 \\

Natural Light Attack
& 94 ($\downarrow 3$)
& 97 ($\downarrow 1$)
& 97 ($\downarrow 1$)
& 93 (--) \\

Shadow Attack
& 84 ($\downarrow 13$)
& 82 ($\downarrow 16$)
& 95 ($\downarrow 3$)
& 79 ($\downarrow 14$) \\

ITA 
& 46 ($\downarrow 51$)
& 64 ($\downarrow 34$)
& 84 ($\downarrow 14$)
& 51 ($\downarrow \textbf{42}$) \\

\textbf{Ours}
& 35 ($\downarrow \textbf{62}$)
& 42 ($\downarrow \textbf{56}$)
& 58 ($\downarrow \textbf{40}$)
& 54 ($\downarrow \ 39$) \\
\bottomrule
\end{tabular}
}
\end{table*}

\subsection{Stage 2: Joint Weather Optimization}

Building upon the fact that Stage 1 has pushed the sample closer to the semantic decision boundary, Stage 2 further introduces multi-scale raindrop and illumination perturbations. It uses a gradient-free optimization algorithm to optimize for more expressive perturbation combinations in the high-dimensional physical parameter space. The final synthesized image $I^{(2)}$ is expressed as:
\begin{equation}
I^{(2)}=I_{r}^{(2)}\odot G
\end{equation}
Here, $I_{r}^{(2)}$ denotes the intermediate rain-perturbed image obtained after Stage 2 raindrop modeling.

To achieve stable and controllable semantic flipping, we introduce an attack margin constraint in Stage 2, requiring the erroneous semantic to exceed the real semantic by at least a given margin:

\begin{equation}
\mathcal{L}_{atk}^{(2)}
= \max\bigl(0,\,
  \delta - \bigl(
    \max_{k\neq y}S_k(I^{(2)}) - S_y(I^{(2)})
  \bigr)
\bigr)
\end{equation}

In terms of perceptual consistency, Stage 2 still explicitly incorporates the VGG perceptual constraint, but adopts a Top-K selective constraint strategy to improve optimization efficiency. Specifically, in each iteration of the gradient-free optimize, candidate solutions are first ranked based on the attack objective and low-level real perceptual constraints. The VGG perceptual loss ~\cite{ref23} is then computed only for the Top-K candidates with the highest attack potential:
\begin{equation}
\mathcal{L}_{perc}^{(2)}=\frac{1}{K}\sum_{j\in\mathrm{Top-}K}\sum_{\ell\in\mathcal{L}}\left\|\phi_\ell(I^{(2,j)})-\phi_\ell(I)\right\|_2^2
\end{equation}
Weighted by $\lambda_{perc}^{(2)}$, this design preserves visual naturalness without prematurely constraining the high-dimensional parameter space.


To further enforce visual realism at the image-structure level, we introduce a regularization term based on structural similarity (SSIM) ~\cite{refjia}, defined as:
\begin{equation}
\mathcal{L}_{\mathit{SSIM}} = 1 - 
\frac{(2\mu_{I^{(2)}}\mu_{I} + C_1)(2\sigma_{I^{(2)}I} + C_2)}
{(\mu_{I^{(2)}}^2 + \mu_{I}^2 + C_1)(\sigma_{I^{(2)}}^2 + \sigma_{I}^2 + C_2)}
\end{equation}
where $\mu$, $\sigma^2$, and $\sigma_{I^{(2)}I}$ denote local means, variances, and covariance, respectively, and $C_1, C_2$ are constants. This term penalizes structural deviations and complements the feature-level perceptual constraint, with strength governed by $\lambda_{\text{real}}^{(2)}$.

The overall optimization objective for Stage 2 is as follows:
\begin{equation}
\begin{aligned}[t]
\mathcal{L}_{stage2} =
\mathcal{L}_{atk}^{(2)} + \lambda_{perc}^{(2)}\mathcal{L}_{perc}^{(2)}
+ \lambda_{real}^{(2)}\mathcal{L}_{SSIM}+ \\
 \lambda_{light}^{(2)}(\bar{G}-G_{0})^{2}
+ \lambda_{range}^{(2)}\max(0,G_{max}-G_{thr})
\end{aligned}
\end{equation}


Owing to the high-dimensional, non-convex raindrop and illumination parameter space with unstable gradients, we adopt CMA-ES for joint black-box optimization. This model-agnostic method is detailed in Appendix A.

\begin{figure*}[htbp]
    \centering
    \includegraphics[width=1\textwidth]{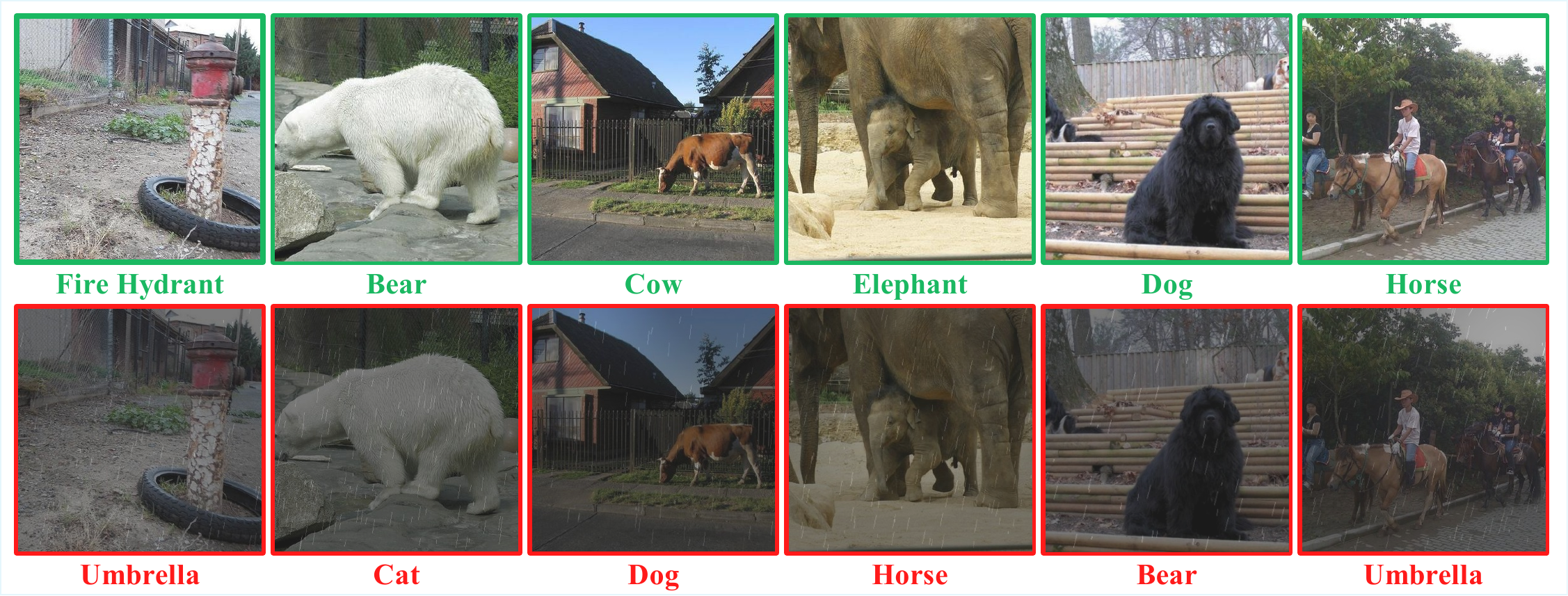}
    \caption{Clean and adversarial samples under weather perturbations. The top row shows clean images with correct labels, while the bottom row shows weather-corrupted adversarial samples with misclassifications.}
    \label{fig3}
    \vspace{-0.3cm}
\end{figure*}

\begin{table*}[h]
\caption{Comparison of attack results (\%) on image captioning tasks.}
\centering
\footnotesize
\renewcommand{\arraystretch}{1.2}

\resizebox{\textwidth}{!}{

\begin{tabular}{c c c c c c c c}
\toprule
Image Encoder & Models & Params & Clean & Natural Light & Shadow & ITA &\textbf{Ours} \\
\midrule

\multirow{4}{*}{OpenAI CLIP ViT-L/14}
& LLaVA-1.5 & 7B & 78.60 & 77.00 ($\downarrow 1.63$) & 74.60 ($\downarrow 4.00$) & 63.73 ($\downarrow 14.87$) & \textbf{56.14 ($\downarrow 22.46$)} \\
& LLaVA-1.6 & 7B & 72.10 & 71.70 ($\downarrow 0.39$) & 71.17 ($\downarrow 0.93$) & 61.60 ($\downarrow 10.51$) & \textbf{46.58 ($\downarrow 25.52$)} \\
& OpenFlamingo & 3B & 70.20 & 69.53 ($\downarrow 0.67$) & 67.80 ($\downarrow 2.40$) & 53.93 ($\downarrow 16.27$) & \textbf{46.10 ($\downarrow 24.10$)} \\
& BLIP-2 (FlanT5$_{\text{XL}}$ ViT-L) & 3.4B & 75.10 & 70.77 ($\downarrow 4.33$) & 68.57 ($\downarrow 6.53$) & 60.93 ($\downarrow 14.17$) & \textbf{51.77 ($\downarrow 23.33$)} \\

\midrule
\multirow{2}{*}{EVA-CLIP ViT-G/14}
& BLIP-2 (FlanT5$_{\text{XL}}$) & 4.1B & 74.96 & 71.27 ($\downarrow 3.69$) & 68.80 ($\downarrow 6.17$) & 62.01 ($\downarrow 11.88$) & \textbf{45.33 ($\downarrow 29.63$)} \\
& InstructBLIP (FlanT5$_{\text{XL}}$) & 4.1B & 76.50 & 72.07 ($\downarrow 4.43$) & 69.77 ($\downarrow 6.73$) & 63.20 ($\downarrow 13.30$) & \textbf{42.86 ($\downarrow 33.64$)} \\

\bottomrule
\end{tabular}
}

\label{tab:2}
\end{table*}

\begin{figure*}[h]
    \centering
    \includegraphics[width=1\textwidth]{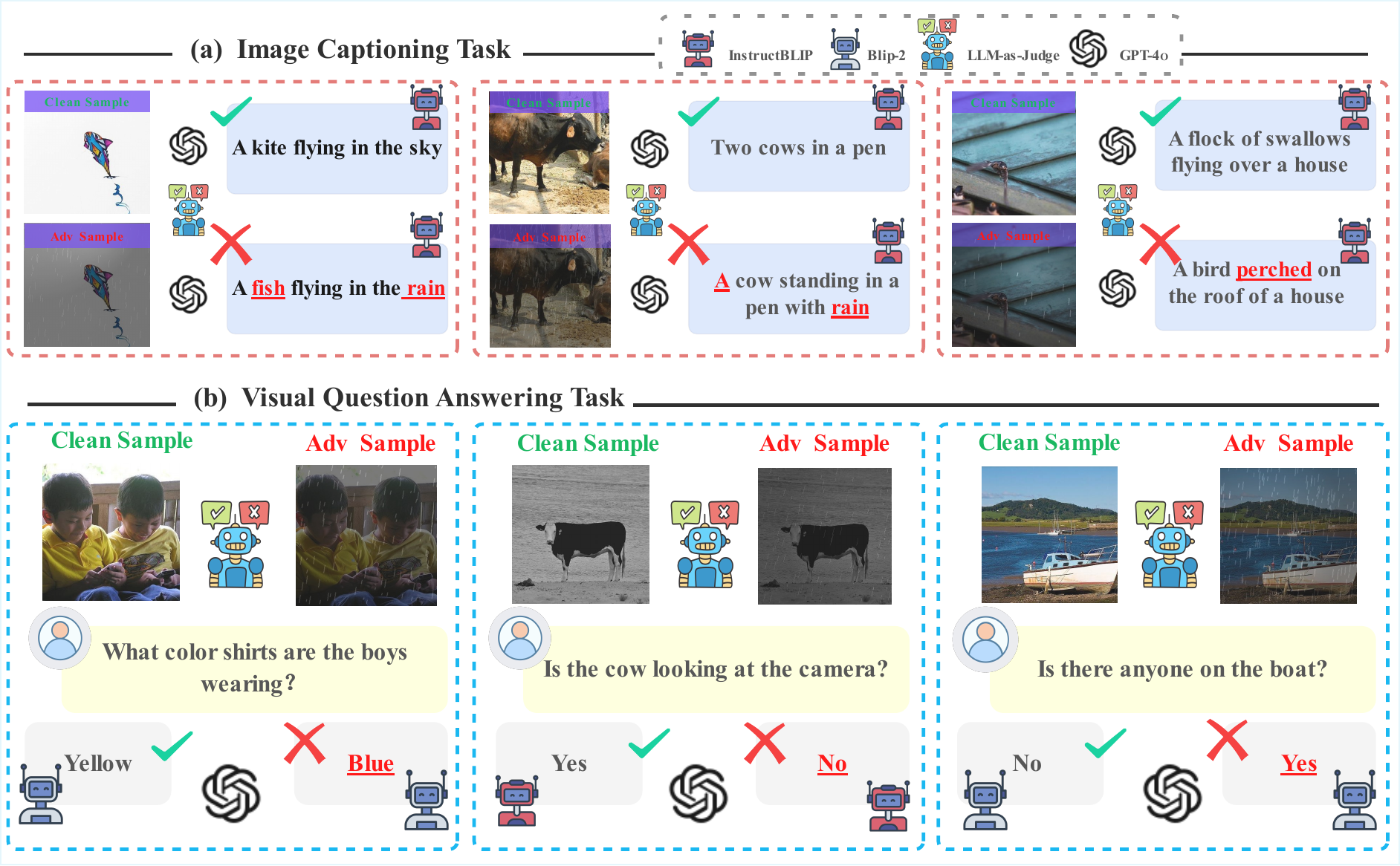}
    \caption{(a) Examples of adversarial sample attacks on the image captioning task; (b) Examples of adversarial sample attacks on the VQA task.}
    \label{fig4}
    \vspace{-0.3cm}
\end{figure*}

\begin{table*}[h]
\caption{Visual question answering task results (\%).}
\centering
\footnotesize
\renewcommand{\arraystretch}{1.2}

\resizebox{\textwidth}{!}{
\begin{tabular}{c c c c c c c c}
\toprule
Image Encoder & Models & Params & Clean & Natural Light & Shadow & ITA & \textbf{Ours} \\
\midrule
\multirow{4}{*}{OpenAI CLIP ViT-L/14}
& LLaVA-1.5 & 7B & 68.00 & 68.00 (-) & 67.00 ($\downarrow 1.00$) & 48.00 ($\downarrow 20.00$) & \textbf{38.00 ($\downarrow 30.00$)} \\
& LLaVA-1.6 & 7B & 64.00 & 63.00 ($\downarrow 1.00$) & 64.00 (-) & 43.00 ($\downarrow 21.00$) & \textbf{37.00 ($\downarrow 27.00$)} \\
& OpenFlamingo & 3B & 45.00 & 39.00 ($\downarrow 6.00$) & 44.00 ($\downarrow 1.00$) & 19.00 ($\downarrow 26.00$) & \textbf{14.00 ($\downarrow 31.00$)} \\
& BLIP-2 (FlanT5$_{\text{XL}}$ ViT-L) & 3.4B & 63.00 & 58.00 ($\downarrow 5.00$) & 50.00 ($\downarrow 13.00$) & 38.00 ($\downarrow 25.00$) & \textbf{30.00 ($\downarrow 33.00$)} \\

\midrule
\multirow{2}{*}{EVA-CLIP ViT-G/14}
& BLIP-2 (FlanT5$_{\text{XL}}$) & 4.1B & 54.00 & 53.00 ($\downarrow 1.00$) & 54.00 (-) & 33.00 ($\downarrow 21.00$) & \textbf{22.00 ($\downarrow 32.00$)} \\
& InstructBLIP (FlanT5$_{\text{XL}}$) & 4.1B & 68.00 & 64.00 ($\downarrow 4.00$) & 62.00 ($\downarrow 6.00$) & 44.00 ($\downarrow 24.00$) & \textbf{38.00 ($\downarrow 30.00$)} \\

\bottomrule
\end{tabular}
}

 \vspace{-0.3cm}
\label{tab:3}
\end{table*}

\section{Experiments}

\subsection{Experimental Settings}

\textbf{Datasets.}\quad We follow the evaluation protocol of ITA~\cite{ref12} and use the same test set of 300 MS COCO images~\cite{ref24} from 30 semantic categories. These categories serve as labels for image classification, with category descriptions constructed using standard prompt templates~\cite{ref15,ref16}. For image captioning and VQA, we adopt identical visual and textual inputs as ITA to ensure fair comparisons. Detailed dataset statistics and category definitions are provided in the Appendix B.1.

\noindent\textbf{Threat Models.}\quad To ensure a fair and direct comparison with ITA, we adopt the same experimental protocol across all tasks.
For zero-shot image classification, we evaluate multiple representative CLIP variants, including OpenCLIP ViT-B/16~\cite{ref25,ref26}, 
Meta-CLIP ViT-L/14~\cite{ref27}, EVA-CLIP ViT-G/14~\cite{ref28}, and OpenAI CLIP ViT-L/14~\cite{ref14}.
For image captioning and VQA, we consider several widely used vision–language models: 
LLaVA-1.5~\cite{ref5}, LLaVA-1.6~\cite{ref4}, OpenFlamingo~\cite{ref29}, 
BLIP-2 ViT-L and BLIP-2 FlanT5-XL~\cite{ref2}, and InstructBLIP~\cite{ref5}.

\noindent\textbf{Baseline Methods.}\quad We compare our approach against state-of-the-art physically consistent environmental attacks, including:
Illumination Transformation Attack (ITA)~\cite{ref12}, 
Shadows Attack~\cite{ref30}, 
and Natural Light Attack~\cite{ref31}.

\noindent\textbf{Implementation Details.}\quad In Stage 1, rain structure and illumination are fixed, and perturbations are generated via multi-scale rain layer superposition. Only the rain–image mixing weight $w_{p}$ is optimized within $[0.02, 0.7]$, with a VGG-based perceptual loss imposed as regularization. The resulting $w_{p}$ is then fixed for the subsequent stage. In Stage 2, CMA-ES is employed to jointly optimize the physical parameters of rainy weather, including rain intensity, geometric attributes, orientation, blur, and density, yielding structurally diverse rain patterns. A continuous illumination model based on two-dimensional Gaussian light sources is incorporated, with illumination parameters optimized while the mixing weight $w_{l}$ is fixed to 0.5.

\subsection{Zero-shot Classification Evaluation}

For zero-shot image classification, we report the Top-1 accuracy drop relative to clean inputs, following ITA for fair comparison. As shown in Table~\ref{tab:1}, our method induces larger performance degradation than ITA on OpenCLIP ViT-B/16, Meta-CLIP ViT-L/14, and EVA-CLIP ViT-G/14, suggesting that jointly modeling multi-scale raindrops and illumination more effectively disrupts visual semantic representations. On OpenAI CLIP ViT-L/14, ITA slightly outperforms our method, likely due to its stronger robustness to local structural perturbations. Figure~\ref{fig3} presents qualitative attack examples, while Appendix B.2 analyzes the misclassification distributions across object categories.

\subsection{Image Captioning Evaluation}

Our method is highly effective on the image captioning task, even though adversarial samples are directly reused from zero-shot image classification without task-specific optimization. We employ GPT-4o as an LLM-as-Judge to assess caption semantic consistency, following prior work, with prompt templates provided in Appendix C. As shown in Table~\ref{tab:2} and Figure~\ref {fig4}(a), our approach induces significantly larger performance drops than ITA under the cross-task transfer setting, demonstrating stronger generalization across both smaller (e.g., OpenFlamingo, BLIP-2) and larger models (e.g., LLaVA-1.5/1.6). These results indicate that multi-scale raindrops and non-uniform illumination cause stable high-level semantic shifts, leading to systematic multimodal misalignment.

\subsection{Visual Question Answering Evaluation}

For VQA, we evaluate our method under weather perturbations using the same cross-task transfer setting, where adversarial examples from zero-shot image classification are reused without additional optimization. We employ GPT-4o as an LLM-as-Judge to assess answer correctness and reasoning consistency (see Appendix D). As shown in Table~\ref{tab:3} and Figure~\ref{fig4}(b), our method induces a significantly larger accuracy drop than ITA. Unlike image captioning, errors primarily arise from incorrect multimodal reasoning, and this effect persists even in large models such as LLaVA-1.5/1.6, revealing fundamental robustness limitations.

\begin{figure*}[h]
    \centering
    \begin{minipage}[t]{0.48\linewidth}
        \centering
        \includegraphics[width=\linewidth]{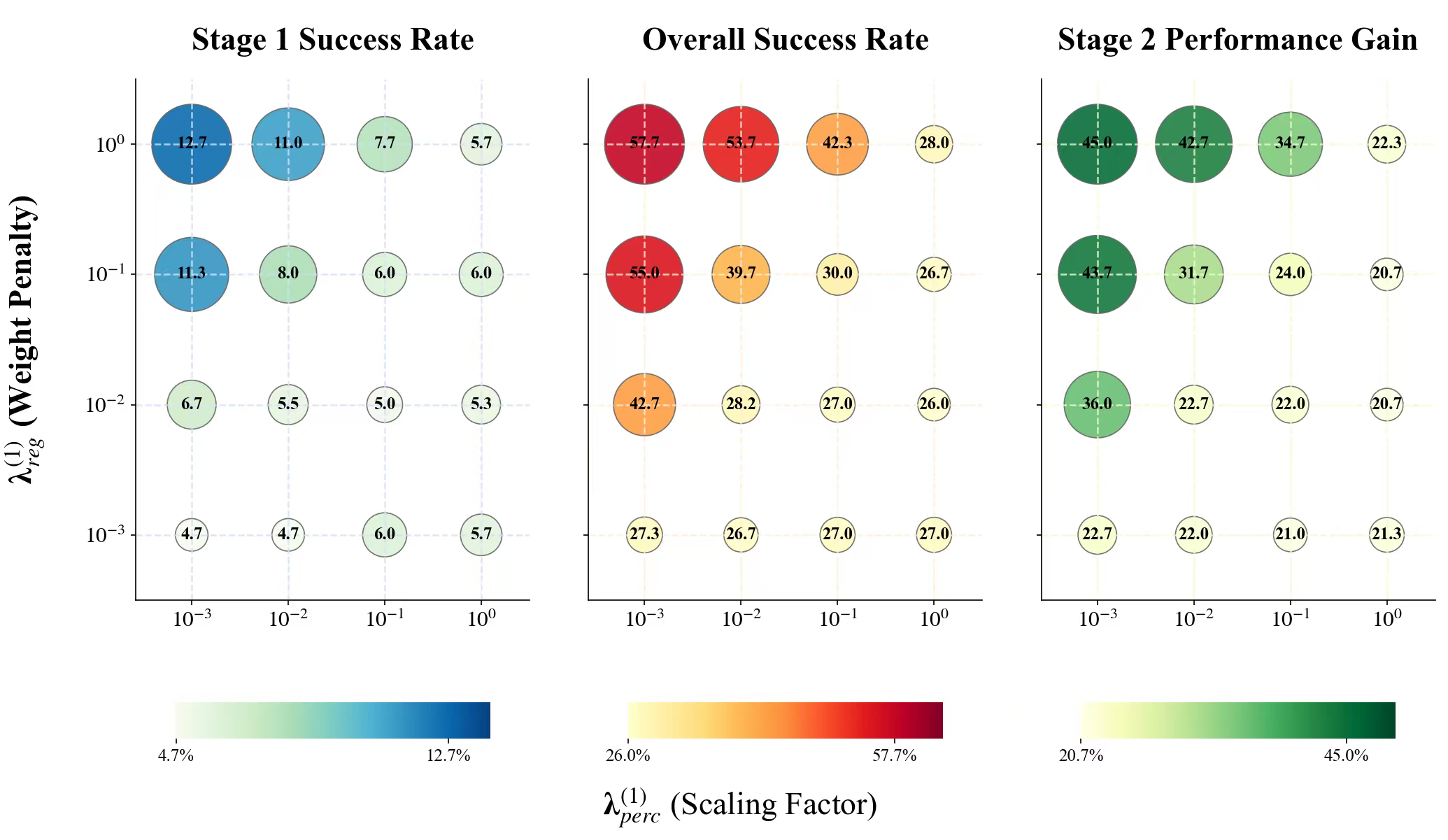}
        \caption{Stage 1 hyperparameter ablation results.}
        \label{fig5}
    \end{minipage}
    \hfill
    \begin{minipage}[t]{0.48\linewidth}
        \centering
        \includegraphics[width=\linewidth]{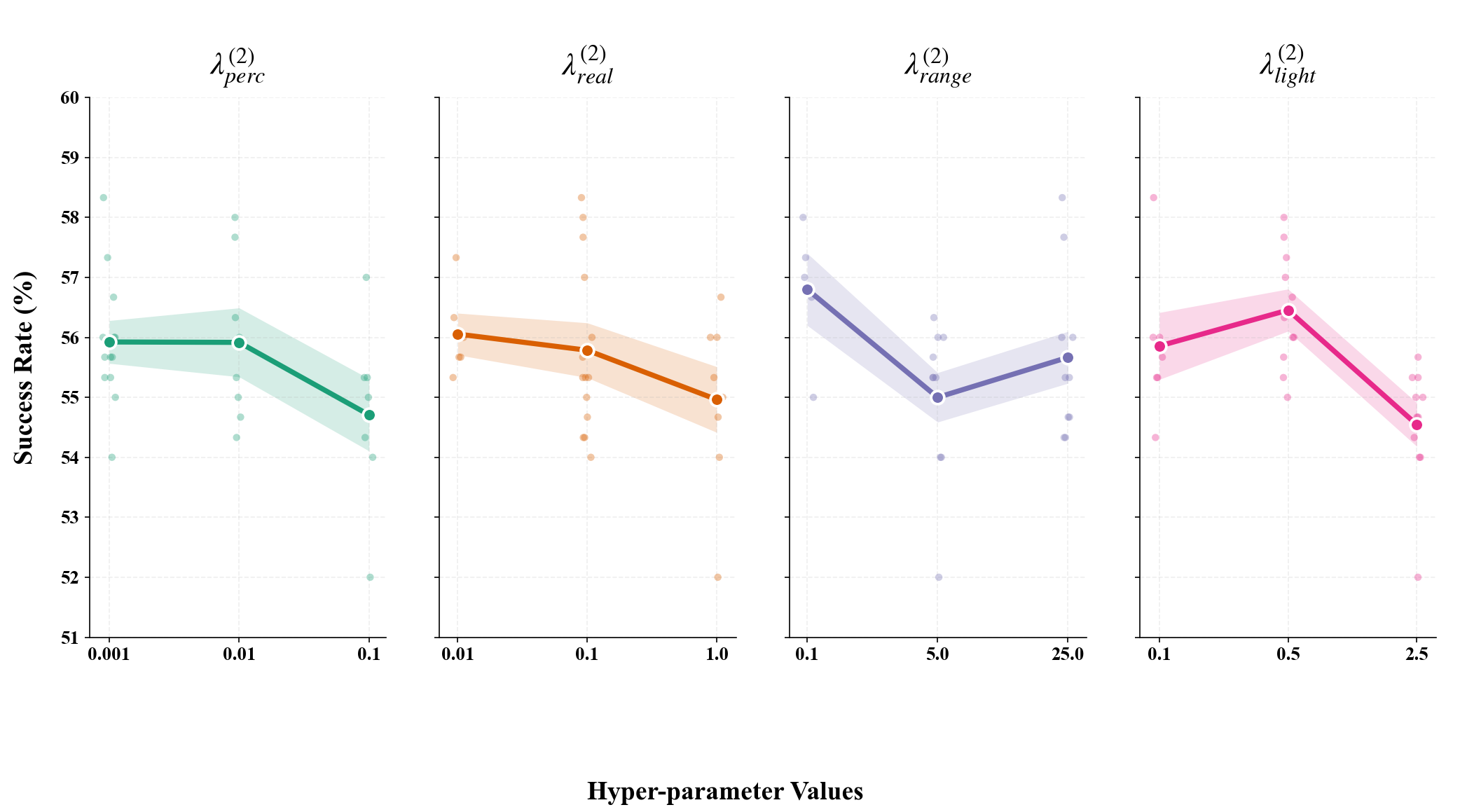}
        \caption{Stage 2 hyperparameter ablation results.}
        \label{fig6}
    \end{minipage}
    \vspace{-0.3cm}
\end{figure*}

\section{Ablation Study}

\textbf{Ablation on Stage 1 Hyperparameters.}\quad We conduct an ablation study on two key hyperparameters in Stage 1: the scaling factor of the regularization on the global raindrop mixing weight $\lambda_{reg}^{(1)}$ and the scaling factor of the perceptual constraint $\lambda_{perc}^{(1)}$, to evaluate their influence on attack performance and subsequent optimization. The experimental results are presented in Figure~\ref{fig5}. For Meta-CLIP ViT-L/14, when $\lambda_{reg}^{(1)}$ is large (e.g., $10^{0}$ or $10^{-1}$) and $\lambda_{perc}^{(1)}$ is small (e.g., $10^{-3}$ or $10^{-2}$), Stage 1 achieves the highest attack success rate of up to 12.7\%. This indicates that moderate rain intensity constraints combined with relaxed perceptual consistency enable faster erosion of decision boundaries. In contrast, excessively large $\lambda_{perc}^{(1)}$ limits perturbation freedom, reducing attack success. Notably, the Stage 1 success rate does not fully determine the final attack outcome, as overall success can still range from 26\% to 57.7\% even when Stage 1 performance is low. These findings demonstrate that Stage 1 mainly provides favorable initialization for Stage 2, substantially improving the effectiveness of CMA-ES optimization. Furthermore, Stage 2 achieves the greatest performance gain when $\lambda_{reg}^{(1)}$ is large and $\lambda_{perc}^{(1)}$ is small, reaching 45.0\%, suggesting that looser constraints in Stage 1 yield more exploratory initial solutions that unlock the optimization potential of Stage 2.

\noindent\textbf{Ablation on Stage 2 Hyperparameters.}\quad  We analyze four key Stage-2 hyperparameters: the perceptual consistency weight $\lambda_{perc}^{(2)}$, realism regularization weight $\lambda_{real}^{(2)}$, range constraint weight $\lambda_{range}^{(2)}$, and illumination modulation weight $\lambda_{light}^{(2)}$. These parameters primarily enforce visual naturalness and physical consistency rather than directly boosting attack success. Figure~\ref{fig6} shows that attack success is stable across a wide range of settings, suggesting that Stage 2’s gradient-free optimization is largely insensitive to hyperparameters. Performance is mainly driven by the physical parameter space and Stage-1 semantic initialization, with strong constraints only mildly limiting the optimize space. Overall, Stage 2 balances attack effectiveness and physical consistency without fine-tuning.

\noindent\textbf{Component-level Ablation.}\quad Table~\ref{tab:ablation_clip} shows that removing either multi-scale raindrops or illumination modulation degrades performance, with raindrop removal causing the largest drop and illumination modulation mainly affecting larger models. This confirms their complementary roles in enabling stable semantic flipping.

\begin{table}[htbp]
\centering
\caption{Performance across CLIP backbones (\%).}
\label{tab:ablation_clip}
\resizebox{\columnwidth}{!}{
\begin{tabular}{ccccc}
\toprule
Methods & \makecell{OpenCLIP\\ViT-B/16} & \makecell{Meta-CLIP\\ViT-L/14} & \makecell{EVA-CLIP\\ViT-G/14} & \makecell{OpenAI CLIP\\ViT-L/14} \\
\midrule
\makecell[c]{w/o Multi-scale\\Raindrops} & 12 & 17 & 13 & 20 \\
\makecell[c]{w/o Illumination\\Modulation} & 36 & 45 & 31 & 27 \\ 
\makecell[c]{ITA} & 51 & 34 & 14 & \textbf{42} \\
\makecell[c]{Ours} & \textbf{62} & \textbf{56} & \textbf{40} & 39 \\
\bottomrule
\end{tabular}
}
\vspace{-0.3cm}
\end{table}



\begin{figure}[h]
    \vspace{-0.4cm}
    \centering
    \includegraphics[width=1\linewidth]{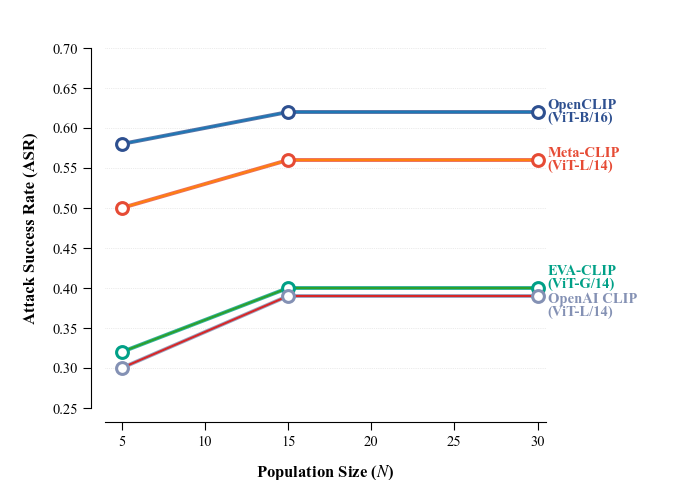} 
    \caption{Impact of population size on optimization.}
    \label{fig7}
    \vspace{-0.5cm}
\end{figure}

\noindent\textbf{Population Size Ablation.}\quad Figure~\ref{fig7} illustrates the effect of population size on attack success across several CLIP-based models. As the population size increases from 5 to 15, performance improves steadily, indicating more effective optimization. However, further increases beyond 15 result in diminishing returns, with performance gains gradually plateauing and showing little improvement beyond that point.

\section{Conclusion}
We propose a semantics-disentangled, two-stage rainy-weather adversarial framework for VLMs. By operating in a physically consistent, non-pixel space, our method produces stable and transferable semantic perturbations across multiple tasks, revealing fundamental robustness limitations under structured weather conditions. Future work will explore more complex environmental factors and corresponding defense strategies.

\section{Limitations}
This study primarily focuses on rainy conditions as a representative weather scenario and does not explicitly cover other complex environmental factors such as fog, haze, or snow. Extending the proposed framework to a broader range of weather conditions would allow for a more comprehensive assessment of vision–language model robustness in diverse real-world environments. In addition, while we evaluate the impact of structured weather perturbations on downstream task performance, there remains room for deeper investigation into finer-grained perturbation mechanisms and the interactions among different environmental factors. A more detailed analysis along these dimensions could further improve the understanding of how complex physical conditions influence cross-modal semantic representations.

\bibliographystyle{acl_natbib}
\bibliography{refs}

\appendix

\section{Algorithmic Details of the Two-Stage Weather Attack}
\label{sec:appendix}

Algorithm \ref{alg:two_stage_attack} provides the complete pseudo-code of our two-stage physically consistent attack. Stage 1 optimizes a low-dimensional global mixing weight to condition the cross-modal embedding space under a perceptual constraint, and Stage 2 applies CMA-ES to optimize the non-differentiable physical parameter space of multi-scale raindrops and illumination to obtain the final adversarial image.

\section{Image classification task}
\label{sec:appendix}
\subsection{Selected COCO Categories}

The 30 categories included in the dataset used for the zero-shot classification task are shown in Table~\ref{tab:coco30}.
\begin{table}[h]
\centering
\small
\captionsetup{position=bottom}

\renewcommand{\arraystretch}{1.08}
\setlength{\tabcolsep}{6pt}

\begin{tabular}{@{} c c @{\hspace{12pt}} c c @{\hspace{12pt}} c c @{}}
\toprule
\textbf{ID} & \textbf{Category} &
\textbf{ID} & \textbf{Category} &
\textbf{ID} & \textbf{Category} \\
\midrule
0  & airplane     & 1  & banana        & 2  & bear \\
3  & bed          & 4  & bird          & 5  & boat \\
6  & broccoli     & 7  & bus           & 8  & cake \\
9  & cell phone   & 10 & clock         & 11 & cow \\
12 & dog          & 13 & donut         & 14 & elephant \\
15 & fire hydrant & 16 & horse         & 17 & kite \\
18 & motorcycle   & 19 & pizza         & 20 & sandwich \\
21 & teddy bear   & 22 & traffic light & 23 & stop sign \\
24 & toilet       & 25 & train         & 26 & umbrella \\
27 & vase         & 28 & zebra         & 29 & sheep \\
\bottomrule
\end{tabular}

\caption{The selected 30 categories in the COCO dataset.}
\label{tab:coco30}
\end{table}

\subsection{Specific Statistics on Attack Categories}
\begin{figure}[htbp]
    \centering
    \includegraphics[width=1\linewidth]{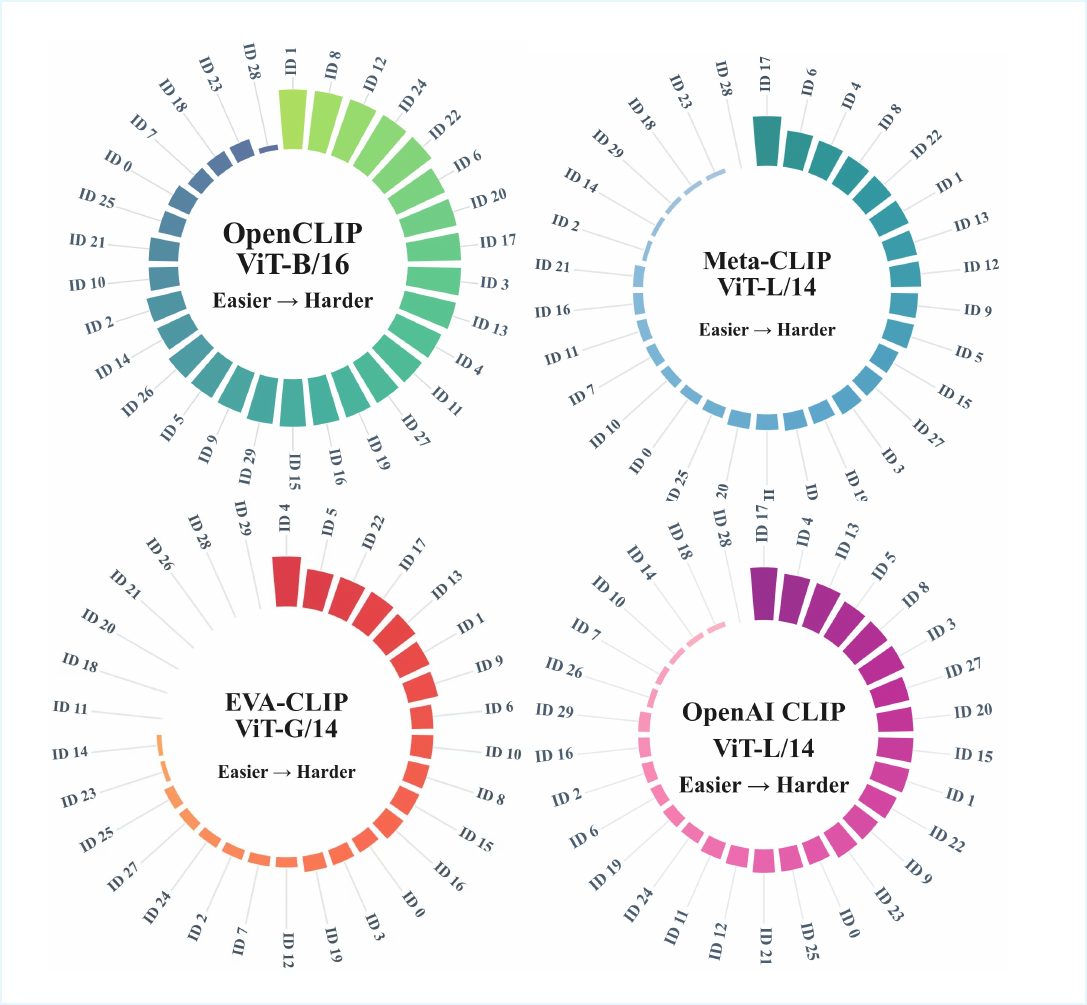}
    \caption{Specific Statistics on Attack Categories.}
    \label{fig666}
    \vspace{-0.3cm}
\end{figure}

The Figure~\ref{fig666} illustrates the specific distribution of classification errors across the four models tested in the image classification task.

\section{LLM-as-Judge Prompts for Image Captioning}
\label{sec:appendix}

\begin{figure}[htbp]
    \centering
    \includegraphics[width=1\linewidth]{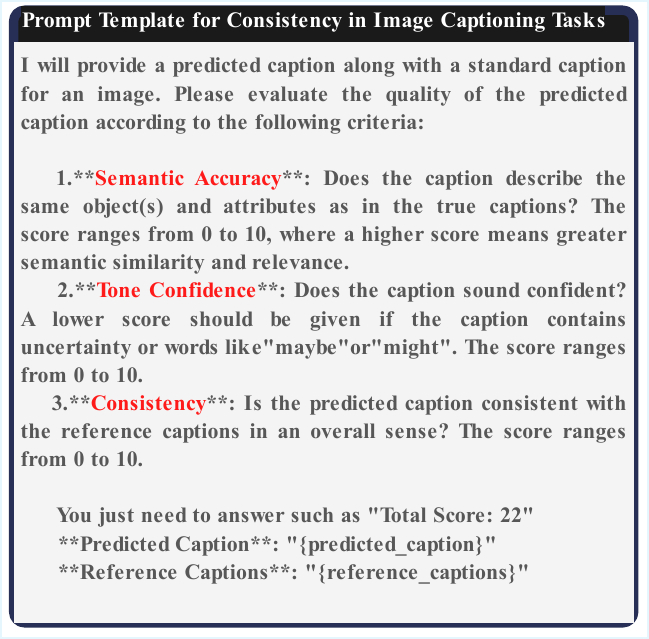}
    \caption{Prompt Template for Consistency in Image Captioning Tasks.}
    \label{fig8}
\end{figure}

Figure~\ref{fig8} presents the prompt template used for LLM-based evaluation of image captioning consistency. Given a predicted caption and the reference caption(s), the judge scores semantic accuracy, tone confidence, and overall consistency, and returns a standardized total score for reproducible comparison.


\section{LLM-as-Judge Prompts for Visual Question Answering}
\label{sec:appendix}
\begin{figure}[htbp]
    \centering
    \includegraphics[width=1\linewidth]{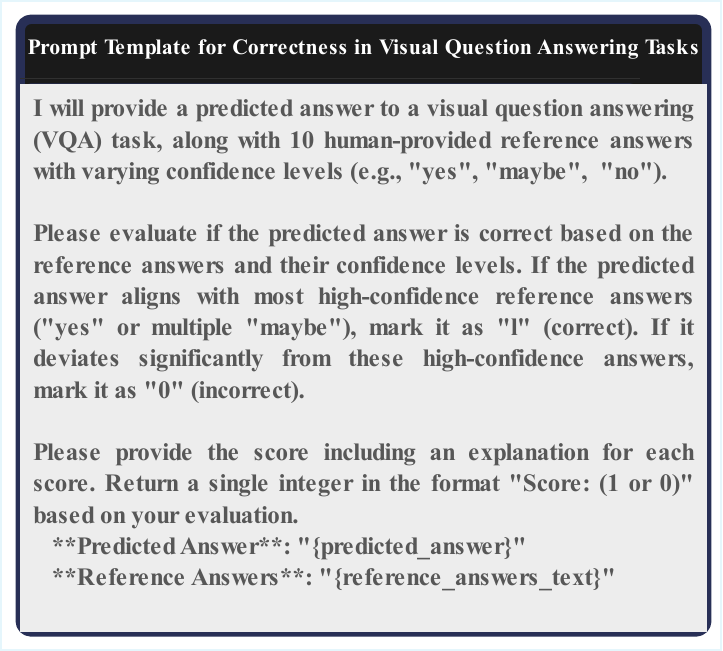}
    \caption{Prompt Template for Consistency in Image Captioning Tasks.}
    \label{fig9}
\end{figure}

Figure~\ref{fig9} shows the prompt template used for LLM-based correctness evaluation in VQA. Given a predicted answer and the set of human reference answers, the judge outputs a binary score (1/0) indicating whether the prediction agrees with the high-confidence references under a standardized format.

\begin{algorithm}[htbp]
\caption{Two-Stage Physically Consistent Adversarial Attack}
\label{alg:two_stage_attack}
\begin{algorithmic}[1]

\Require Original image $I$; visual encoder $f_v$; text encoder $f_t$;
semantic prompts $\{T_k\}$; Stage-1 iterations $T_1$; Stage-2 iterations $T_2$
\Ensure Adversarial image $I_{\mathrm{adv}}$

\Statex
\State \textbf{Stage 1: Global Semantic Perturbation}
\State Initialize global rain layer $R$ with fixed spatial structure
\State Initialize mixing coefficient $\alpha \gets 0$
\For{$t = 1$ to $T_1$}
    \State Generate intermediate image
    $I_t \gets (1 - \alpha) I + \alpha R$
    \State Extract visual embedding
    $z_v \gets f_v(I_t)$
    \State Extract text embedding
    $z_t \gets f_t(T_k)$
    \State Compute semantic margin loss
    $\mathcal{L}_{\mathrm{atk}}^{(1)}$
    \State Compute perceptual consistency loss
    $\mathcal{L}_{\mathrm{perc}}^{(1)}$
    \State Compute overall Stage-1 objective:
    \[
        \mathcal{L}_{\mathrm{stage1}} =
        \mathcal{L}_{\mathrm{atk}}^{(1)}
        + \lambda_{\mathrm{perc}}^{(1)} \mathcal{L}_{\mathrm{perc}}^{(1)}
        + \lambda_{\mathrm{reg}}^{(1)} \mathcal{L}_{\mathrm{reg}}
    \]
    \State Update $\alpha$ by minimizing
    $\mathcal{L}_{\mathrm{stage1}}$
\EndFor
\State Obtain intermediate image $I_{\mathrm{stage1}}$

\Statex
\State \textbf{Stage 2: Physically Consistent Optimization}
\State Initialize multi-scale rain parameters $\Theta_r$
\State Initialize illumination parameters $\Theta_\ell$ (gain map $G$)
\State Initialize CMA-ES optimize distribution
\For{$t = 1$ to $T_2$}
    \State Sample candidate parameters
    $\{\Theta_r^{(i)}, \Theta_\ell^{(i)}\}$
    \State Generate physically consistent candidate images
    \State Evaluate semantic margin and realism constraints
    \State Rank candidates based on attack objective
    \State Select Top-$K$ candidates
    \State Compute perceptual loss
    $\mathcal{L}_{\mathrm{perc}}^{(2)}$ for Top-$K$ candidates
    \State Update CMA-ES distribution
\EndFor
\State \Return $I_{\mathrm{adv}}$

\end{algorithmic}
\end{algorithm}

\end{document}